\documentclass[conference,10pt]{IEEEtran}
\IEEEoverridecommandlockouts
\usepackage{amsmath,amssymb,amsfonts}
\usepackage{adjustbox}
\usepackage{algorithmic}
\usepackage{threeparttable}
\usepackage{array}
\usepackage{fontawesome}
\usepackage{graphicx}
\usepackage{multirow}
\usepackage{textcomp}
\usepackage{enumitem}
\usepackage[english]{babel}
\usepackage[utf8]{inputenc}
\usepackage[nolist]{acronym}
\usepackage{amsmath}											
\usepackage{amsfonts}											
\usepackage{blindtext} 											
\usepackage{booktabs}											
\usepackage{fancyvrb}
\usepackage{ifthen}												
\usepackage{filecontents} 										
\usepackage[T1]{fontenc} 
\usepackage{graphicx}
\usepackage[dvipsnames,table]{xcolor}
\usepackage{ifthen}												
\usepackage{listings}											
\usepackage{lipsum}
\usepackage{multirow}
\usepackage{pgfplots}
\usepackage{siunitx}
\usepackage{ragged2e}						
\usepackage{tabularx}
\usepackage{url}
\usepackage{tikz}
\usepackage{circuitikz}
\usepackage{comment}
\usepackage{footnote}
\usepackage{float}
\usepackage{listings}
\usepackage{setspace}
\usepackage{makecell, tabularx}
\usepackage{enumitem}

\usetikzlibrary{calc,trees,positioning,arrows,chains,shapes.geometric,%
	decorations.pathreplacing,decorations.pathmorphing,shapes,%
	matrix,shapes.symbols}

\usepackage[font=footnotesize,labelfont=bf]{subcaption}		
\usepackage{caption}
\usepackage[ruled,linesnumbered]{algorithm2e}
\SetAlCapNameFnt{\small}
\SetAlCapFnt{\small}
\SetAlgoNlRelativeSize{0}
\RestyleAlgo{ruled}
\SetKwComment{Comment}{/* }{ */}

\usepackage[style=ieee, backend=biber, natbib=true, maxbibnames=1, minbibnames=1]{biblatex}
\usepackage{wrapfig}
\usepackage{fancyhdr}

\fancypagestyle{firstpage}{%
    \fancyhf{}%
    \fancyhead[C]{\normalsize To appear at the IEEE International Conference on Intelligent Processing, Hardware, Electronics, and Radio Systems (CIPHER), February 13-15, 2026, NIT Jalandhar, India}
    
    \fancyfoot[C]{\footnotesize \copyright2026 IEEE.  Personal use of this material is permitted.  Permission from IEEE must be obtained for all other uses, in any current or future media, including reprinting/republishing this material for advertising or promotional purposes, creating new collective works, for resale or redistribution to servers or lists, or reuse of any copyrighted component of this work in other works.}
}

\lstdefinelanguage{Verilog}{morekeywords={accept_on,alias,always,always_comb,always_ff,always_latch,and,assert,assign,assume,automatic,before,begin,bind,bins,binsof,bit,break,buf,bufif0,bufif1,byte,case,casex,casez,cell,chandle,checker,class,clocking,cmos,config,const,constraint,context,continue,cover,covergroup,coverpoint,cross,deassign,default,defparam,design,disable,dist,do,edge,else,end,endcase,endchecker,endclass,endclocking,endconfig,endfunction,endgenerate,endgroup,endinterface,endmodule,endpackage,endprimitive,endprogram,endproperty,endspecify,endsequence,endtable,endtask,enum,event,eventually,expect,export,extends,extern,final,first_match,for,force,foreach,forever,fork,forkjoin,function,generate,genvar,global,highz0,highz1,if,iff,ifnone,ignore_bins,illegal_bins,implements,implies,import,incdir,include,initial,inout,input,inside,instance,int,integer,interconnect,interface,intersect,join,join_any,join_none,large,let,liblist,library,local,localparam,logic,longint,macromodule,matches,medium,modport,module,nand,negedge,nettype,new,nexttime,nmos,nor,noshowcancelled,not,notif0,notif1,null,or,output,package,packed,parameter,pmos,posedge,primitive,priority,program,property,protected,pull0,pull1,pulldown,pullup,pulsestyle_ondetect,pulsestyle_onevent,pure,rand,randc,randcase,randsequence,rcmos,real,realtime,ref,reg,reject_on,release,repeat,restrict,return,rnmos,rpmos,rtran,rtranif0,rtranif1,s_always,s_eventually,s_nexttime,s_until,s_until_with,scalared,sequence,shortint,shortreal,showcancelled,signed,small,soft,solve,specify,specparam,static,string,strong,strong0,strong1,struct,super,supply0,supply1,sync_accept_on,sync_reject_on,table,tagged,task,this,throughout,time,timeprecision,timeunit,tran,tranif0,tranif1,tri,tri0,tri1,triand,trior,trireg,type,typedef,union,unique,unique0,unsigned,until,until_with,untyped,use,uwire,var,vectored,virtual,void,wait,wait_order,wand,weak,weak0,weak1,while,wildcard,wire,with,within,wor,xnor,xor,`uvm_create, `uvm_rand_send_with},morecomment=[l]{//}}

\newboolean{blindreview}
\setboolean{blindreview}{false}
\DeclareRobustCommand{\IEEEauthorrefmark}[1]{\smash{\textsuperscript{\footnotesize #1}}}

\usepackage[left=1.57cm,right=1.57cm,top=1.75cm,bottom=2.5cm,letterpaper]{geometry}

\definecolor{yamlKey}{rgb}{0.01,0.49,0.33}   
\definecolor{yamlString}{rgb}{0.6,0.0,0.0} 
\definecolor{yamlComment}{rgb}{0.4,0.4,0.4}

\usepackage{listings}
\usepackage{xcolor}
\lstdefinelanguage{YAML}{
	keywords={system_prompt,goal,coverage_hole_analyzer,sva_property_generator},
	keywordstyle=\color{yamlKey}\bfseries,
	comment=[l]\#, 
	commentstyle=\color{yamlGray}\ttfamily,
	stringstyle=\color{yamlStr},
	morestring=[b]',
	morestring=[b]",
	literate=
	{true}{{{\color{yamlBool}true}}}{4}
	{false}{{{\color{yamlBool}false}}}{5}
	{:}{{{\color{yamlGray}:}}}{1}
	{-}{{{\color{yamlGray}-}}}{1}
	{>}{{{\color{yamlGray}>}}}{1}
}

\lstdefinelanguage{JSON}{
	string=[s]{"}{"},
	stringstyle=\color{blue},
	breakatwhitespace=false,
	showspaces=false,                
	showstringspaces=false,
	comment=[l]{:},
	commentstyle=\color{black},
	literate=
	{true}{{{\color{yamlBool}true}}}{4}
	{false}{{{\color{yamlBool}false}}}{5}
	{:}{{{\color{yamlGray}:}}}{1}
	{-}{{{\color{yamlGray}-}}}{1}
	{>}{{{\color{yamlGray}>}}}{1}
}

\begin{document}	
	
\begin{acronym}[]
    
    \acro{AI}[AI]{Artificial Intelligence}
    \acro{ADHD}[ADHD]{Attention Deficit Hyperactivity Disorder}
    \acro{ASIC}[ASIC]{Application Specific Integrated Circuit}
    \acro{AGI}[AGI]{Artificial General Intelligence}
    \acro{API}{Application Programming Interface}
    \acro{CEX}[CEX]{Counter Example}
    \acrodefplural{CEX}[CEXs]{Counter Examples}
    \acro{CoT}[CoT]{Chain-of-Thought}
    \acro{DUV}[DUV]{Design Under Verification}
    \acro{DOS}[DOS]{Denial of Service}
    \acro{DSGI}[DSGI]{Domain-Specific General Intelligence}
    \acro{EDA}[EDA]{Electronic Design Automation}
    \acro{FV}[FV]{Formal Verification}
    \acro{GenAI}[GenAI]{Generative AI}
    \acro{GPT}[GPT]{Generative Pre-trained Transformer}
    \acro{HIL}[HIL]{Human-in-the-Loop}
    \acro{HDL}[HDL]{Hardware Description Language}
    \acro{IP}[IP]{Intellectual Property}
    \acrodefplural{IP}[IPs]{Intellectual Properties}
    \acro{IC}[IC]{Integrated Chip}
    \acro{KPI}[KPI]{Key Performance Indicator}
    \acrodefplural{KPI}[KPIs]{Key Performance Indicators}
    \acro{KG}{Knowledge Graph}
    \acrodefplural{KG}[KGs]{Knowledge Graphs}
    \acro{LLM}[LLM]{Large Language Model}
    \acrodefplural{LLM}[LLMs]{Large Language Models}
    \acro{ML}[ML]{Machine Learning}
    \acro{MAS}[MAS]{Multi-Agent System}
    \acro{NLP}[NLP]{Natural Language Processing}
    \acro{RTL}[RTL]{Register Transfer Level}
    \acro{RAG}[RAG]{Retrieval Augmented Generation}
    \acro{RQ}[RQ]{Research Question}
    \acrodefplural{RQs}[RQs]{Research Questions}
    \acro{RDF}{Resource Description Framework}
    \acro{SoC}[SoC]{System-on-Chip}
    \acrodefplural{SoC}[SoCs]{System-on-Chips}
    \acro{SVA}[SVA]{SystemVerilog Assertion}
    \acrodefplural{SVA}[SVAs]{SystemVerilog Assertions}
    \acro{STSC}[STSC]{Short Term, Short Context}
    \acro{UVM}[UVM]{Universal Verification Methodology}
    \acro{vPlan}[vPlan]{Verification Plan}
    \acro{VLSI}[VLSI]{Very Large Scale Integration}
\end{acronym}


\title{Agentic AI-based Coverage Closure for Formal Verification}

\ifthenelse{\boolean{blindreview}}{}{
	\author{
		\IEEEauthorblockN{
	        Sivaram Pothireddypalli\IEEEauthorrefmark{1},
            Ashish Raman\IEEEauthorrefmark{2},
            Deepak Narayan Gadde\IEEEauthorrefmark{3},
            Aman Kumar\IEEEauthorrefmark{1}}
         \vspace{0.3cm}   
		\IEEEauthorblockA{
			\IEEEauthorrefmark{1}Infineon Technologies India Private Limited, India\\
            \IEEEauthorrefmark{2}Dr. B R Ambedkar National Institute of Technology Jalandhar, India\\
            \IEEEauthorrefmark{3}Infineon Technologies Dresden AG \& Co. KG, Germany \\}
		}
}

\maketitle

\thispagestyle{firstpage}

\begin{abstract}
Coverage closure is a critical requirement in \ac{IC} development process and key metric for verification sign-off. However, traditional exhaustive approaches often fail to achieve full coverage within project timelines. This study presents an agentic AI-driven workflow that utilizes \ac{LLM}-enabled \acf{GenAI} to automate coverage analysis for formal verification, identify coverage gaps, and generate the required formal properties. The framework accelerates verification efficiency by systematically addressing coverage holes. Benchmarking open-source and internal designs reveals a measurable increase in coverage metrics, with improvements correlated to the complexity of the design. Comparative analysis validates the effectiveness of this approach. These results highlight the potential of agentic AI-based techniques to improve formal verification productivity and support comprehensive coverage closure.

\end{abstract}


\section{Introduction} \label{sec:introduction}

Hardware design verification plays a vital role in the \ac{VLSI} development cycle, ensuring that the implemented design accurately meets its intended specifications. The most widely used methods in industry are simulation-based verification and formal methods. In simulation-based verification, input stimuli are applied to the design, and the resulting outputs are analyzed to verify functional correctness. This method takes  significant time to time to verify \ac{RTL} designs and also risks omitting corner cases \cite{BSIStudy875Summary}. Formal methods are an alternative to simulation, in which \ac{RTL} designs are checked for all possible input cases using mathematical modeling. In a formal environment, there are various tasks, such as planning, property development, proof execution, and formal sign-off. During the property development phase, SystemVerilog assertions are used to verify the design's functional correctness. In addition, it has the potential to simplify the verification process and improve the overall productivity \cite{Kern1999Survey}.

The entire process of verification even using the traditional exhaustive methods of study takes up to \SI{60}{\percent} of the project time \cite{VerStudy}. The primary feature of the effort is the realization of functional coverage because the coverage is often used to measure the progress in verification. It shows that the pre-determined verification
plan is to be implemented and satisfied.

Achieving formal coverage closure within a set project timeline is a persistent challenge in hardware design verification. To overcome the above challenge, this study explores the application of \ac{LLM}-enabled \ac{GenAI}, which demonstrates contextual reasoning and can translate natural-language descriptions into executable artifacts. This approach includes context-sensitive query solutions and automated code generation \cite{unknown}, \cite{deepseek}. Both formal specification documents and design artifacts, including \ac{RTL} code, cover properties and assertions that are inherently textual. Therefore, they are well-suited for language-based processing by \acp{LLM}.

Recent advances in agentic AI frameworks have automated key aspects of hardware design and verification. These frameworks can interpret natural language specifications, generate \ac{RTL} code, and perform design verification. One automatic assertion-generation framework uses \acp{LLM} to generate assertions by analyzing natural-language specifications for each architectural signal. This framework incorporates three customized \acp{LLM} to extract structural specifications, map signal definitions, and generate assertions. This approach provides substantial advantages for preventing bugs during design and detecting bugs during verification \cite{assertllm}. Another framework that uses a Multi-Agent System (MAS) to support collaborative code generation, iterative \ac{RTL} refinement, and dynamic property generation for language-based processing by \acp{LLM}. In such an automated environment, specialized agents are capable of generating \ac{RTL} codes and interfacing with sophisticated \ac{EDA} tools for validation and verification \cite{gadde}. Similarly, Saarthi (ScalAble ARTificial and Human Intelligence) is a fully autonomous \ac{AI} formal verification engineer designed to verify \ac{RTL} designs end-to-end through an agentic workflow \cite{saarthi}. Saarthi uses agentic reasoning patterns in conjunction with human supervision to achieve robust verification results. Starting with the initial design specification, it sequentially handles verification planning, assertion generation, property proving, \acp{CEX} analysis, and formal coverage evaluation to support verification sign-off.

Even with these automated assertion generation and verification execution runs, existing workflows frequently result in residual coverage gaps. This work introduces an agentic AI method, building on Saarthi, to address the coverage gaps.
 
The main contributions of this paper are as follows:

\begin{itemize}
\item \textit{Automated gap classification}: Analyzes coverage report generated by a formal verification tool to partition uncovered \ac{RTL} regions into control flow branches (\textit{case}, \textit{if/else}) and statement level gaps.
\item \textit{Targeted property generation}: An \ac{LLM} generates SystemVerilog properties (assertions and covers) using \ac{RTL} and specification context to verify the intended behavior for each uncovered location.
\item \textit{Iterative agentic AI coverage closure}: New properties are incorporated, and the coverage report is re-evaluated in a loop until predefined thresholds are met.
\end{itemize}

This method precisely targets uncovered \ac{RTL} code sections and automatically generates SystemVerilog properties. As a result, coverage gaps are effectively reduced, and the overall verification effort required to achieve the desired coverage within a defined timeframe is minimized.

\section{Background} \label{sec:background}
This section introduces agentic AI patterns and their ability to handle complex tasks, followed by an overview of coverage principles within the context of formal verification.

\subsection{Agentic Workflows}
The implementation of \ac{MAS}, which interacts and communicates among several special software agents to accomplish complicated tasks, is known as agentic AI. These agents usually utilize \acp{LLM}, and they have tool-awareness to perform the task. Task scheduling, task allocation, and establishment of feedback are parts of workflow to enhance the quality of the response. A logging mechanism guarantees the possibility to generate all the system processes throughout the working process. Further, \ac{HIL} checkpoints: these are checkpoints that provide monitoring of the model drift, output hallucinations, and workflow decision making where these are deemed essential. 

The major benefits of the agentic patterns are as follows:
\begin{itemize}
\item \textit{Dynamic Task Decomposition}: Dynamic breakdown of a complex and multi-stage problem to a set of simpler sub problems thus maximizing scalability and execution efficiency  \cite{jeyakumar2024advancing}.
\item \textit{Deliberation loops make performance better}: Built-in propose-critique-repair cycles in the system inherently limit hallucinations, certify in-between results, and hence provide evaluative feedback of refinements, leading to much better performance  \cite{chen2025}.
\item \textit{Self-Refinement Mechanisms}: Structured self-reflection greatly improves problem-solving accuracy across models and benchmarks \cite{Self-ref}.
\item \textit{Multi-Agent workflow}:  Multi-objectives are broken down to form smaller tasks and the allocation to agents is determined by their expertise which makes them superior in reasoning and also makes the system stable \cite{MAS}.
\item \textit{Orchestration}: It dictates the way tasks are carried out, it regulates the sharing of data between agents and also dictates dependencies between agents that are interdependent. This kind of coordination can provide a high level of reliability in operations and avoid possible conflicts in workflow definitions \cite{Orch}.
\item \textit{Invocation of external tools or APIs}: Agentic AI enables the use of outside tools and APIs to perform tangible activities and judge their effectiveness as opposed to making use of language-based reasoning only. This feedback process enhances precision and reliability, thereby increasing autonomous operation in the course of complex operational processes \cite{TOOLS}.

\end{itemize}

 The characteristics given above increase the independence and capability of an agentic AI to solve difficult tasks. Such systems in formal verification speed up the signoff verification process. The processes involved are to produce the verification plan based on the natural-language specification, produce SystemVerilog properties and systematically fills if coverage gaps are present. As the flow goes on, the critic agent checks the property correctness, whereas the executor agent communicates with the formal verification tool to establish the properties and review the \acp{CEX}. Therefore, it offers a scalable verification framework of a complex \ac{RTL} code, correspondence with the design intent, and quicken the process to formal verification.

\subsection{Formal Coverage}
Formal coverage quantifies the extent to which verification objectives are rigorously exercised within a given set of environmental assumptions. The most common coverage metric is checker coverage, including line, expression, and toggle coverage. Line coverage specifically reports the percentage of synthesizable \ac{RTL} statements within the \ac{DUV} that have been mathematically proven reachable and subsequently exercised by a specified property set within the formal verification environment.
\definecolor{yamlGray}{gray}{0.35}
\definecolor{codeBG}{RGB}{248,248,248}
 
\begin{lstlisting}[language=verilog,
numberstyle=\scriptsize\color{gray}\ttfamily,
numbersep=12pt,
xleftmargin=2em,
numbers=left,
backgroundcolor=\color{codeBG},
captionpos=t,
rulecolor=\color{yamlGray}, basicstyle=\ttfamily\scriptsize,
caption={Consider the following RTL block},
label={lst:RTL}] 
always @(posedge clk) begin
 if (a && b)
  c <= d1;
 else
  c <= d2;
end
\end{lstlisting}

\begin{lstlisting}[language=verilog,
numberstyle=\scriptsize\color{gray}\ttfamily,
numbersep=12pt,
xleftmargin=2em,
numbers=left,
backgroundcolor=\color{codeBG},
captionpos=t,
rulecolor=\color{yamlGray}, basicstyle=\ttfamily\scriptsize,
caption={SVA property exercising the line 3 of RTL block},
label={lst:RTL_out}]
 property branch_captures_d1;
  @(posedge clk)
   (a && b) |=> (c == $past(d1));
 endproperty
 assert property (branch_captures_d1);
\end{lstlisting}

Consider a minimal example that yields two distinct line coverage targets, corresponding to lines 3 and 5, as detailed in Listing~\ref{lst:RTL}. Listing~\ref{lst:RTL_out} illustrates a formal property that, when \textbf{(a \&\& b)} holds at a rising clock edge, checks that \textbf{c} captures the prior value of \textbf{d1} in \textbf{n} cycles. This specific statement is marked as covered once the formal tool successfully identifies a witnessing trace or inductively proves the implication holds universally. If the formal environment is over-constrained (thereby preventing the antecedent \textbf{(a \&\& b)} from ever being satisfied) or if the proof/search depth parameters are insufficient, the target statement may remain unexercised, and coverage remains incomplete. If a comprehensive test suite fails to exercise line 5, the aggregate line coverage for that specific suite is reported as \SI{50}{\percent}. Consequently, line coverage metrics provide a direct indication of whether a specific coverage target has been successfully exercised or not. The systematic incorporation of targeted assertions and cover properties for previously uncovered branches of \ac{RTL} code can substantially improve coverage in formal verification \cite{seligman2015formal}.

\section{Methodology} \label{sec:methodology}
This section outlines the methodology for achieving coverage closure in formal verification using agentic \ac{AI}. First, an overview of the agentic AI-driven workflow for formal verification is presented. Subsequently, the systematic approach employed to ensure comprehensive coverage is discussed.

\subsection{Overview of Saarthi Workflow}
\begin{figure}[h!]
\centering
  \includegraphics [width=0.50\textwidth] {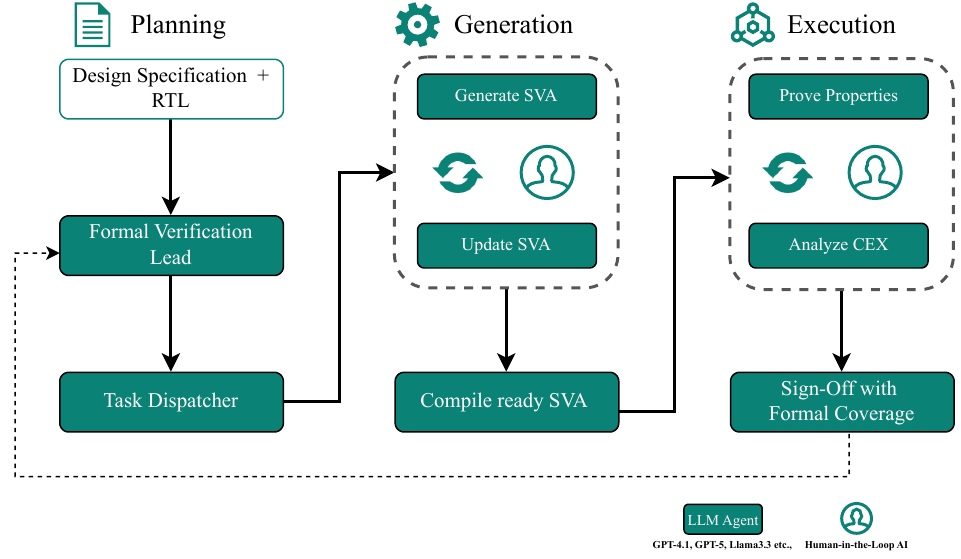}
\caption{Agentic AI methodology for formal verification \cite{saarthi}}
\label{fig:saarthi}
\end{figure}

Saarthi uses a sophisticated agentic \ac{AI}-based workflow that makes use of multi-agent collaboration to enable formal verification, as shown in the Fig.~\ref{fig:saarthi}. The formal verification lead agent starts the process by parsing the available \ac{RTL} and the design specification to create a structured \ac{vPlan}. Specialized verification agents follow project conventions and coding templates while translating each \ac{vPlan} item into compile-ready \acp{SVA} under the direction of a task dispatcher. Critic agents give specific feedback to help properties get better while always checking the generated \acp{SVA} for syntax errors, completeness, and semantic accuracy. As the flow progress toward high-quality, tool-acceptable \acp{SVA}, the iterative loop represented in the planning and generation stages, ensures alignment with the initial requirements. Before proceeding further, Saarthi refers the task to human experts for disambiguation if predetermined iteration thresholds are not met without resolution.

 In order to prove properties, gather \acp{CEX}, and facilitate automated repair cycles, the code executor agent interfaces with industry-standard formal verification tools such as Cadence Jasper Gold \cite{cadenceJasper} during execution. Agents for verification and criticism examine \ac{CEX} traces to identify underlying issues, revise \acp{SVA} or \ac{RTL} constraints as necessary, and submit proofs again until convergence. The following section presents a detailed explanation of how coverage agents incorporate additional properties when coverage gaps remain, ensuring alignment with \ac{vPlan} targets.

\begin{figure}[h!]
\centering
  \includegraphics [width=0.5\textwidth] {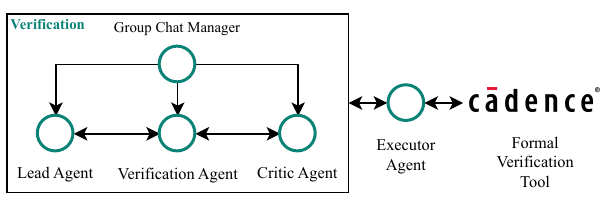}
\caption{Depiction of agentic \ac{AI} performing workflow execution \cite{gadde}}
\label{fig:MAS_flow}
\end{figure}

The suggested framework, as shown in the Fig.~\ref{fig:MAS_flow}, uses a multi-agent group chat where specialized roles, such as \textit{verification lead}, \textit{formal verification engineer}, and \textit{SystemVerilog expert}, work together on verification tasks defined by the user through a shared event-driven mechanism. A \ac{HIL} agent continuously supervises the workflow to ensure that automated decisions are consistent with expert domain knowledge. In an event-driven architecture, a \textit{Group Chat Manager} uses AutoGen's Core API to make sure that only one agent can talk at a time and that everyone follows the rules of interaction \cite{autogen2024}. To get around some of the common problems with \acp{LLM}, complicated goals are broken down into smaller tasks that are given to different agents. At the same time, critic agents check the quality and accuracy of the \acp{SVA} that are made over and over again. A feedback loop that can only go through five times causes humans to step in when progress stops, which cuts down on hallucination and keeps cycles from being unproductive. By combining automated agent collaboration with strategic human oversight, this architecture creates a scalable, auditable, and high-confidence verification workflow for next-generation \ac{IC} design.
 
\subsection{Agentic {AI} based coverage closure}

\begin{figure}[h!]
\centering
  \includegraphics [width=0.45\textwidth] {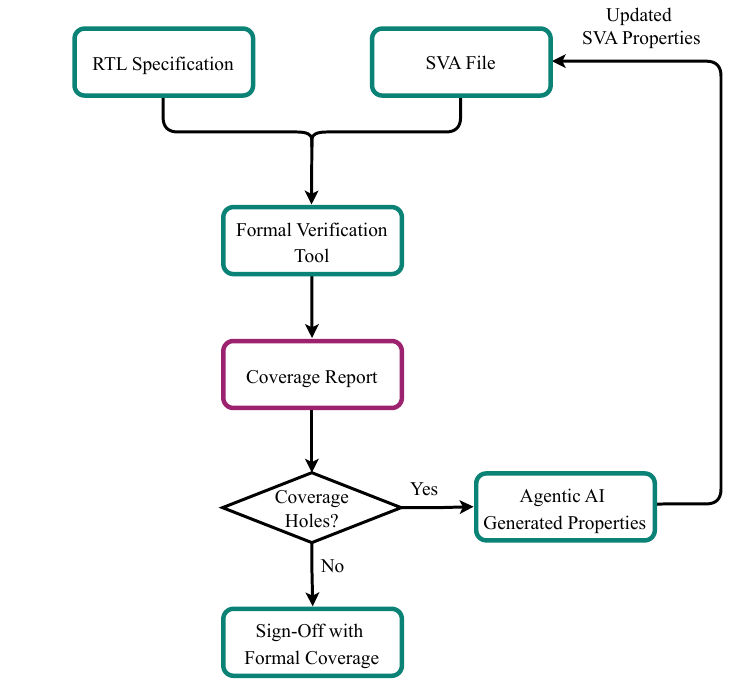}
\caption{Basic flow diagram of coverage  pipeline}
\label{fig:Flow}
\end{figure}

The methodology begins with the \ac{RTL} specification and an initial \ac{SVA} file, which are processed by a formal verification tool to generate a coverage report after proof process. If no coverage holes, such as uncovered or unreachable sections of the \ac{RTL} codebase, are discovered, the design proceeds directly to formal sign-off, as illustrated in Fig. \ref{fig:Flow}. When gaps are present, the procedure transfers control to an agentic \ac{AI} stage that creates \ac{LLM}-generated properties aimed at the uncovered areas in order to make ready for iterative closure in the next stages.

The agentic \ac{AI} methodology addresses coverage gaps in formal verification by ingesting the coverage report generated by a formal verification tool and automatically producing a modified \ac{SVA} file. This process is orchestrated by two cooperating agents: (i) a \textit{Coverage Hole Analyzer} that characterizes the \ac{RTL} context of uncovered code and (ii) an \textit{SVA Property Generator} that generates targeted, \ac{LLM}-generated properties to address the identified gaps. The updated \ac{SVA} properties are incorporated into the \ac{SVA} file and then reintroduced to the formal verification tool, thereby closing the feedback loop. The iterative process continues until all required coverage goals are fully met.

\begin{figure}[h!]
\centering
  \includegraphics [width=0.45\textwidth] {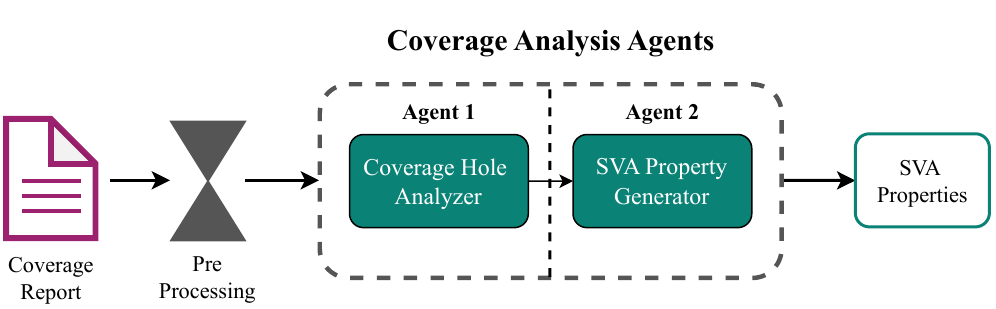}
\caption{Overview of the coverage  agent workflow}
\label{fig:Overview}
\end{figure}

Coverage analysis data is obtained from the formal verification tool. Following the required preprocessing and normalization steps, exact \ac{RTL} locations of uncovered regions are extracted for each module, including any preconditions identified by the tool. These locations are then classified into two groups: (i) isolated locations, representing standalone sections with minimal structural context, and (ii) branch and statement locations, which are related to control flow, such as \textit{case}, \textit{if/else}, and \textit{always} blocks. This classification ensures that attributes are properly targeted for coverage closure by influencing downstream agents' reasoning regarding reachability and timing considerations. The pre\-processed locations are input to the \textit{Coverage Hole Analyzer}, and its output is subsequently provided to the \textit{SVA Property Generator}, which generates the required \ac{SVA} properties, as shown in Fig.~\ref{fig:Overview}.

The \textit{Coverage Hole Analyzer} agent is a core component of our methodology that systematically identifies areas within the design that show insufficient coverage. The structural and semantic context surrounding every coverage hole is reconstructed through static code analysis by interpreting pre\-processed locations and their corresponding preconditions with relevant \ac{RTL} files. It extracts the name of the module being analyzed, checks the timing behavior, gets the input and output signals, and maintains records of the preconditions and reset logic for each area that is found. It also picks the type of enclosing block, like \texttt{always\_ff}, \texttt{always\_comb}, \texttt{if/else}, or \texttt{case}, to show how the design operates and the way it fits into the design hierarchy. This detailed context is encoded into a short JSON format, which eliminates the need for extra \ac{RTL} parsing and lets downstream agents easily figure out properties. A sample definition and goal utilized by the \textit{Coverage Hole Analyzer} agent are detailed in Listing~\ref{lst:SVA_prom}, while the corresponding output generated is presented in Listing~\ref{lst:SVA_out}.

\begin{lstlisting}[language=YAML,numberstyle=\scriptsize\color{gray}\ttfamily,backgroundcolor=\color{codeBG},numbersep=12pt,xleftmargin=2em,numbers=left,captionpos=t,rulecolor=\color{yamlGray}, basicstyle=\ttfamily\scriptsize, caption={Concise definition and targets for \textit{Coverage Hole Analyzer}}, label={lst:SVA_prom}]
coverage_hole_analyzer:
 system_prompt: >  You are the RTL coverage hole analyzer with mandatory code-extarction capabilities.

 goal: >  Transform coordinate-based location data into detailed RTL behaviour analysis. Follow the steps below

  STEP 1: RTL CODE EXTARCTION
  -For every location,extract the exact RTL slice

  STEP 2: SINGLE BEHAVIOUR CLASSIFICATION
  -One location -> one behaviour -> one future property.

  STEP 3: LOGIC PATTERN and CONSOLIDATION ANALYSIS
  -For each location,derive a logic_signature:
   {opeartion_pattern, signal_realtion, 
      timing_pattern}

\end{lstlisting}

\begin{lstlisting}[language=JSON, basicstyle=\ttfamily\scriptsize, backgroundcolor=\color{codeBG}, rulecolor=\color{yamlGray},numberstyle=\scriptsize\color{gray}\ttfamily, numbersep=12pt,xleftmargin=2em,captionpos=t,numbers=left, caption={Response generated from \textit{Coverage Hole Analyzer} in JSON format}, label={lst:SVA_out}]
{
  "module": "alu",
  "input_type": "BRANCH_STRUCTURE|ISOLATED_STRUCTURE",
  "locations": [
    {
      "start": [60, 34],
      "end": [60, 39]
    }
  ],
  "type": "BRANCH",
  "code": "c <= a + b",
  "behavior": "Mode-based state machine branch selection,
   "statement_type": "case_statement",
   "signals": {
     "in": ["a", "b"],
     "out": ["c"]
   },
   "timing": "always_ff"
}
\end{lstlisting}

The \textit{SVA Property Generator} processes the JSON output data and extracts macro definitions from the  main \ac{SVA} file to generate \acp{SVA}. It applies templates specific to each construct and timing class to produce SystemVerilog properties that target unexercised branches and assertions that formalize intended behavior under the specified preconditions. Clocking and reset semantics are determined by the timing class, and signals are mapped to the appropriate interfaces. During property generation, the agent reuses an existing property name if it is present in the \ac{SVA} file; otherwise, it generates a random property name to minimize duplication. The agent then merges these properties with the existing properties, maintains naming conventions, and inserts traceability comments linking each property to its original coverage location. The resulting \ac{SVA} file is immediately prepared for re-execution within the formal verification tool environment. A sample definition and goal utilized by the \textit{SVA Property Generator} agent are detailed in Listing~\ref{lst:gen_prom}, while the extracted code from the JSON response of the agent is presented in Listing~\ref{lst:gen_out}.

\begin{lstlisting}[language=YAML,numberstyle=\scriptsize\color{gray}\ttfamily,backgroundcolor=\color{codeBG},numbersep=12pt,xleftmargin=2em,numbers=left,captionpos=t,rulecolor=\color{yamlGray}, basicstyle=\ttfamily\scriptsize, caption={Concise definition and targets for \textit{SVA Property Genertor}}, label={lst:gen_prom}]
sva_property_generator:
 system_prompt: >  You are Expert in generating 
    SystemVerilog properties for uncovered RTL 
    locations based on the JSON output.

  goal: > Transform location analysis into compliant SVA properties using only available signals.

  STEP 1: Resource scan:
     -Extract the signals, macros, and parameters/constants available at the SVA file.

  STEP 2: Single-statement enforcement:
     -Allowed forms are (antecedent) |-> (single_outcome), |=>, or 
    |-> ##N, matching the existing style
      
  STEP 3: Timing style match:
     -For always_ff, prefer |=> if an antecedent exists; otherwise replicate the observed 
      form |->  ##1)
     -For always_comb  or continuous assignments, use |-> (single cycle).
  
\end{lstlisting}

\begin{lstlisting}[language=verilog,
	numberstyle=\scriptsize\color{gray}\ttfamily,
	numbersep=12pt,
	xleftmargin=2em,
	numbers=left,
	backgroundcolor=\color{codeBG},
	captionpos=t,
	rulecolor=\color{yamlGray}, basicstyle=\ttfamily\scriptsize,
	caption={Code extracted from \textit{SVA Property Generator}'s JSON response},label={lst:gen_out}]
	property sum_of_a_and_b;
	@(posedge clk) disable iff (!rst)
	1'b1 |=> c == $past(a + b);
	endproperty
	assert property (sum_of_a_and_b);
\end{lstlisting}

The workflow operates iteratively; after each update to the \ac{SVA} file, the formal verification tool is re-executed to generate a new coverage report. Autonomous agents update or append the formal properties set in response to any coverage gaps until the target coverage metrics are met. When the design intent cannot be verified using purely static analysis, an expert agent gets involved at each iteration to validate property intent, avoid over-constraint of the specification, and resolve complex edge cases. This iterative process thus facilitates the faster closure of coverage, traceably reducing the manual verification effort through systematic consistency throughout the process. 

\section{Benchmarking} \label{sec:results}
To evaluate the effectiveness and capabilities of the proposed methodology for coverage closure in formal verification, we benchmark the updated Saarthi framework, integrated with our coverage agents, on a diverse set of \ac{RTL} designs. The benchmark suite includes both open-source and in-house industry-grade design spanning various complexity levels. Table~\ref{tab:coverage_metrics} highlights performance of the Saarthi framework with and without proposed coverage agents. To compare performance across various \acp{LLM}, we employ two \acp{KPI}. The first is the proven percentage, defined as the percentage of properties successfully proven among all properties generated by the \ac{LLM}; the second is coverage, quantified as the formal coverage achieved following an end-to-end formal verification flow.

\begin{table}[H]
\centering
\caption{Saarthi performance across designs with coverage agents}
\begingroup
\setlength{\tabcolsep}{1.5pt} 
\renewcommand{\arraystretch}{0.85} 
\tiny 
\resizebox{0.95\columnwidth}{!}{%
\begin{tabular}{@{}llccccccc@{}}
\toprule
\multirow{2}{*}{\textbf{Design}} & \multirow{2}{*}{\textbf{Metric}} & \multicolumn{3}{c}\textbf{{\textbf{Without coverage agents}}} & \multicolumn{3}{c}{\textbf{With coverage agents}} \\
\cmidrule(lr){3-5} \cmidrule(lr){6-8}
& &\textbf{GPT-4.1} & \textbf{GPT-5} & \textbf{Llama3.3} & \textbf{GPT-4.1} & \textbf{GPT-5} & \textbf{Llama3.3} \\
\midrule
\multirow{3}{*}{\textbf{ECC}} & \# Properties & 40 & 23 & 15 & 110 & 35 & 41 \\
& Proven (\%) & 90 & 95.65 & 73.33 & 83.36 & 80 & 60.97 \\
& Coverage (\%) & 61.72 & 70.30 & 70.71 & 89.30 & 92.11 & 89.60 \\
\midrule
\multirow{3}{*}{\textbf{CIC Decimator} ~\cite{nvidiacvdp}} & \# Properties & 13 & 28 & 10 & 28 & 33 & 17 \\
& Proven (\%) & 61.5 & 53.57 & 50 & 82.14 & 78.78 & 52.94 \\
& Coverage (\%) & 63.98 & 71.54 & 52.38 & 74.32 & 81.04 & 66.39 \\
\midrule
\multirow{3}{*}{\textbf{AXI4LITE} ~\cite{nvidiacvdp}} & \# Properties & 35 & 38 & 20 & 101 & 80 & 77 \\
& Proven (\%) & 57.14 & 55.26 & 20 & 64.35 & 42.45 & 51.94 \\
& Coverage (\%) & 27.87 & 45.85 & 24.82 & 36.63 & 53.59 & 51.29 \\
\midrule
\multirow{3}{*}{\textbf{Automotive IP}} & \# Properties & 27 & 35 & 2 & 53 & 89 & 19 \\
& Proven (\%) & 74.07 & 62.22 & 50 & 77.35 & 51.68 & 26.31 \\
& Coverage (\%) & 80.48 & 72.67 & 5.81 & 83.04 & 79.05 & 38.87 \\
\midrule
\multirow{3}{*}{\textbf{Memory Scheduler} \cite{nvidiacvdp}} & \# Properties & 28 & 22 & 30 & 53 & 103 & 47 \\
& Proven (\%) & 53.57 & 45.45 & 26.66 & 73.58 & 90.29 & 29.78 \\
& Coverage (\%) & 50.88 & 46.39 & 46.45 & 68.82 & 67.91 & 49.41 \\
\bottomrule
\end{tabular}
}
\endgroup
\label{tab:coverage_metrics}
\end{table}
Across various models, the Saarthi workflow with our coverage agents achieved the highest performance with OpenAI’s GPT-5 \cite{openai2024}, moderate performance with OpenAI’s GPT-4.1 \cite{openai2024}, and the lowest performance with Meta’s Llama3.3 \cite{meta2024llama3_3}. The incorporation of our coverage agents consistently increases the coverage metrics by approximately \SI{10}{\percent} to \SI{20}{\percent}, with the largest gains observed in the more complex designs. Specifically, in the case of the ECC design, coverage surpassed \SI{80}{\percent} when the coverage agents were employed. Also, considerable enhancement in coverage were recorded for the AXI4LITE, CIC Decimator, and Memory Scheduler designs. The most significant change was observed in the Automotive IP, where the baseline coverage without agents was low but increased significantly after the incorporating of coverage agents. Although GPT-5 exhibited superior performance due to its advanced reasoning capabilities, but exhibited higher latency in comparision. 
\begin{figure*}[t]
 \centering
    \captionsetup[subfigure]{%
            font=footnotesize,
            labelformat=parens,
            labelsep=space,
            justification=centering,
            singlelinecheck=false,
            skip=4pt
    }

    \begin{subfigure}[c]{0.32\textwidth} 
        \centering
        \includegraphics[width=\linewidth]{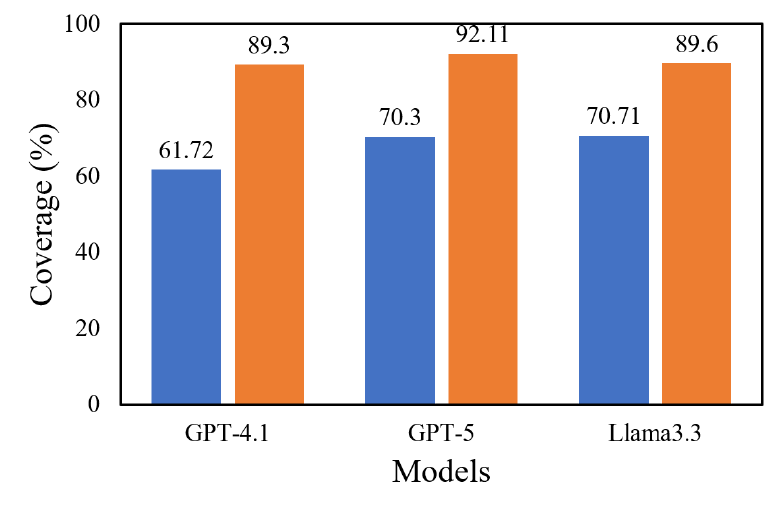}
        \caption{Design: ECC}
        \label{fig:a1}
    \end{subfigure}
    \hfill
    \begin{subfigure}[c]{0.32\textwidth}
        \centering
        \includegraphics[width=\linewidth]{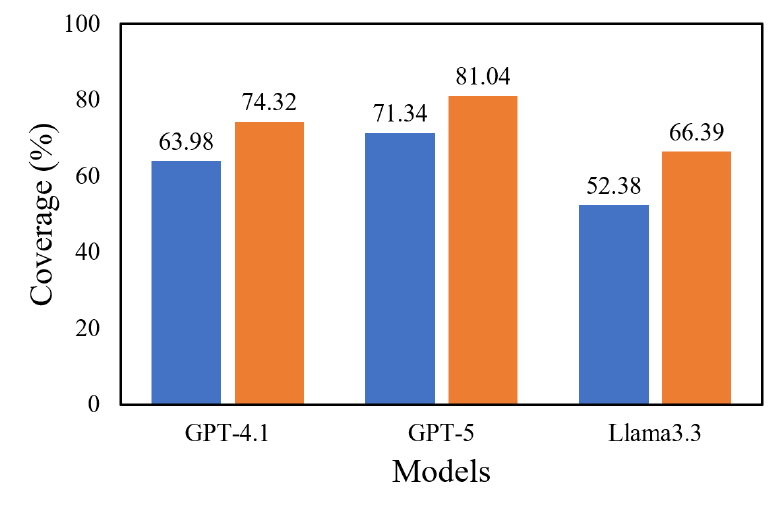}
        \caption{Design: CIC Decimator}
        \label{fig:b1}
    \end{subfigure}
    \hfill
    \begin{subfigure}[c]{0.32\textwidth}
        \centering
        \includegraphics[width=\linewidth]{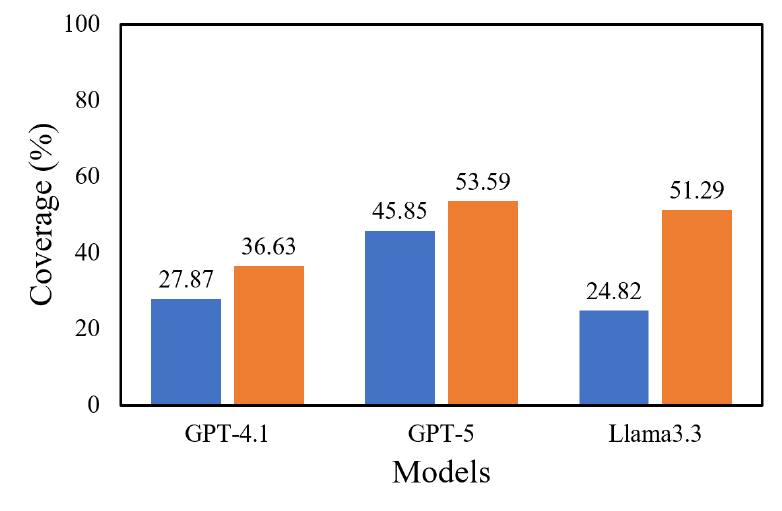}
        \caption{Design: AXI4LITE}
        \label{fig:c1}
    \end{subfigure}

    \vspace{2mm}

    \begin{subfigure}[c]{0.32\textwidth}
        \centering
        \includegraphics[width=\linewidth]{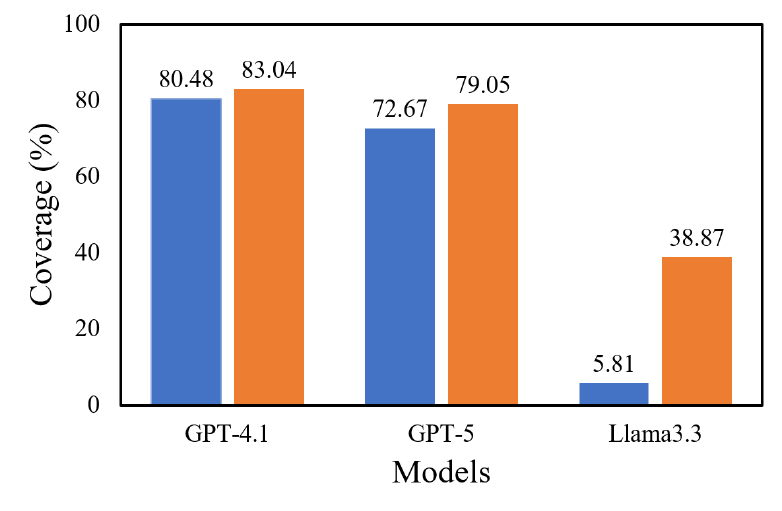}
        \caption{Design: Automotive IP}
        \label{fig:b2}
    \end{subfigure}
    \hfill
    \begin{subfigure}[c]{0.32\textwidth}
        \centering
        \includegraphics[width=\linewidth]{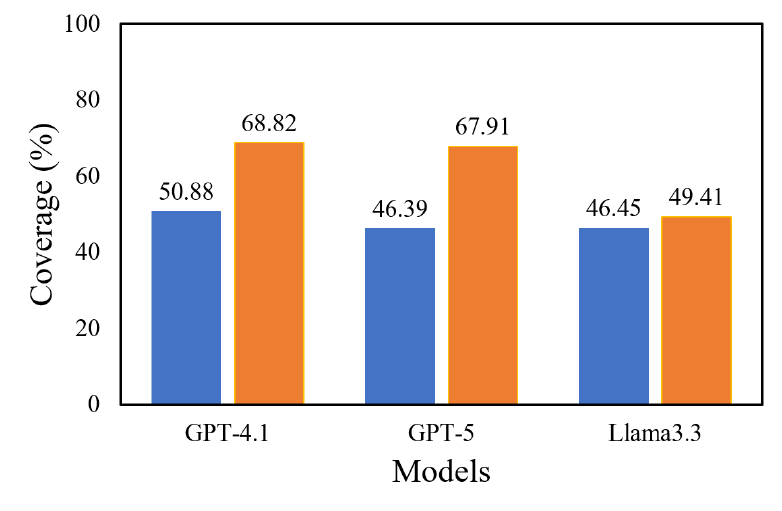}
        \caption{Design: Memory Scheduler}
        \label{fig:c2}
    \end{subfigure}
    \hfill
    \begin{subfigure}[c]{0.32\textwidth}
        \centering
        \includegraphics[width=0.7\linewidth, keepaspectratio]{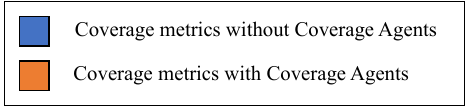}
    \end{subfigure}

    \caption{Coverage metrics comparison across various designs using different models}
    \label{fig:grid}
\end{figure*}

In some cases, especially within a subset of designs, the proven rate for generated properties decreased after the coverage agents’ workflow, even though overall coverage increased. This divergence arises because introducing new properties can enhance coverage, but may also result in formal properties that are more challenging to verify. Furthermore, the effectiveness of the generated properties is closely tied to the extraction of accurate \ac{RTL} code by the \textit{Coverage Hole Analyzer} agent. Imperfections at this stage, along with the occasional generation of incorrect or duplicate properties by \acp{LLM}, can negatively impact the proven rate. The above problems are methodically handled through an \ac{HIL} review stage, guaranteeing that only valid and verifiable properties are retained in the final verification environment.

Importantly, these results were obtained without any \ac{HIL} in the flow. Based on these benchmarking results, integrating \ac{HIL} in the Saarthi flow, specifically after the coverage agents have performed their tasks, is likely to further boost coverage metrics, as new properties have been generated for previously uncovered regions. This enhancement may speed up the formal verification sign-off stage in the \ac{IC} design process and lower the overall time required for completion.

\section{Conclusion} \label{sec:conclusion}
This paper presented an automated workflow that leverages agentic AI enabled \ac{LLM} to accelerate coverage in the formal verification. The proposed approach strategically targets uncovered \ac{RTL} regions, enabling the automatic generation of formal properties. Experimental results from both open-source
and internal \ac{RTL} designs confirm that the framework consistently enhances formal coverage. The target of the approach is formal property generation and the passing rate of generated properties, each further impacting the coverage rate. Even though the proven rate is mostly dependent on the quality of \acp{LLM}' outputs, the findings show that the \ac{LLM}-guided verification is a promising route to automated coverage closure in formal verification. This approach substantially reduces manual effort in the formal verification process and provides a strong methodological basis for future advancements in \ac{AI}-assisted coverage closure.

\section*{Acknowledgement}


\begin{figure}[h!]
\centering
\begin{minipage}{.5\linewidth}
  \centering
  \includegraphics[width=.8\linewidth]{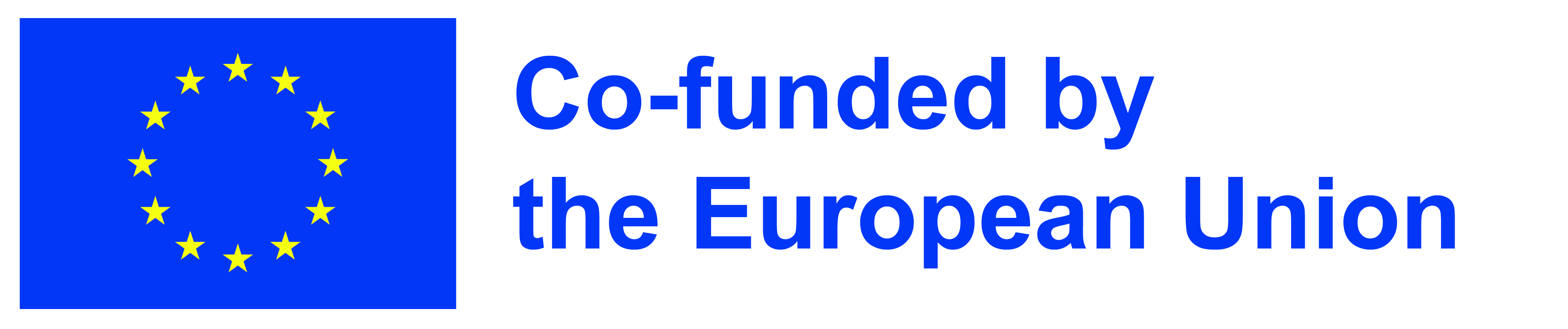}
\end{minipage}%
\begin{minipage}{.5\linewidth}
  \centering
  \includegraphics[width=.6\linewidth]{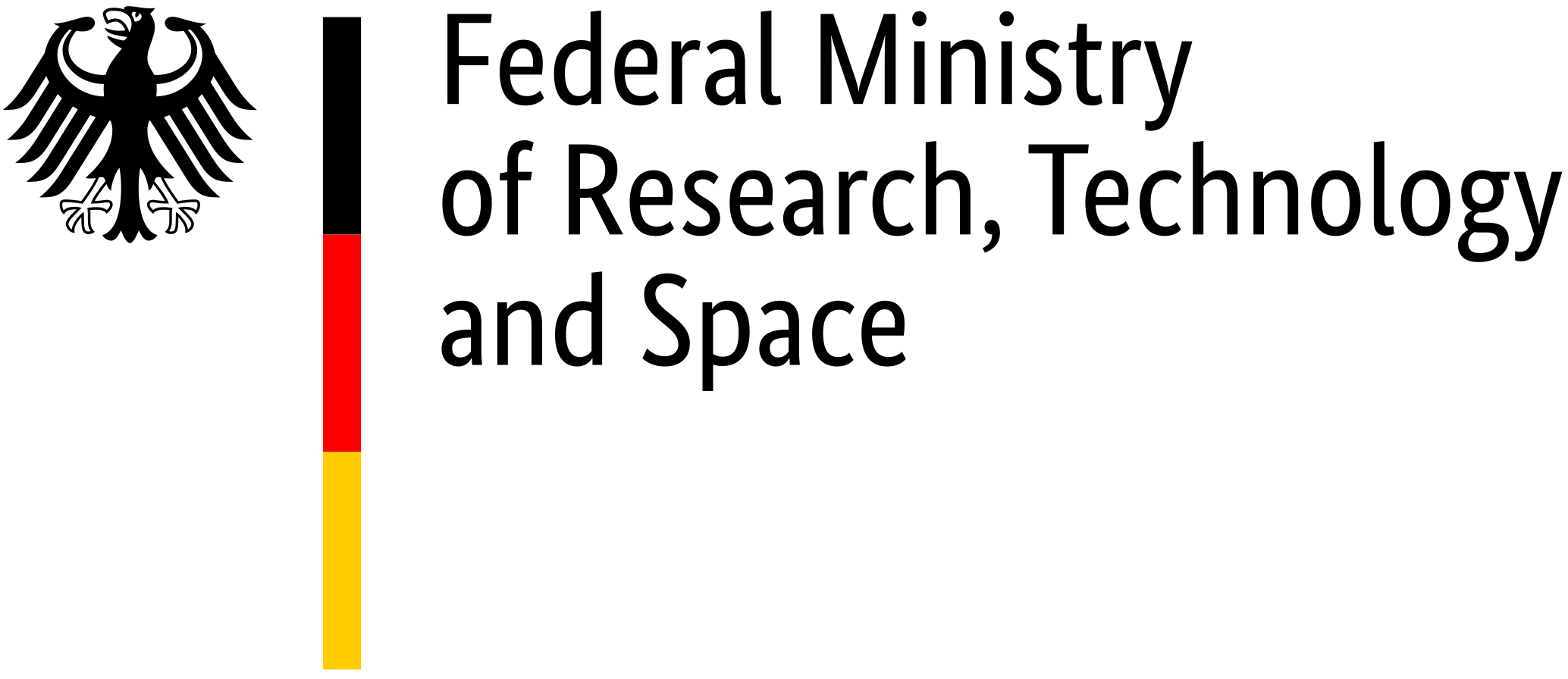}
\end{minipage}
\end{figure}
Under grant 101194371, Rigoletto is supported by the Chips Joint Undertaking and its members, including the top-up funding by the National Funding Authorities from involved countries.

Rigoletto is also funded by the Federal Ministry of Research, Technology and Space under the funding code 16MEE0548S. The responsibility for the content of this publication lies with the author.
\printbibliography 

@techreport{BSIStudy875Summary,
  author    = { A. Leventi-Peetz1},
  title     = {{Formal Methods for Safe and Secure Computer Systems}},
  year={2013},
}

@article{Kern1999Survey,
author = {Kern, Christoph and Greenstreet, Mark R.},
title = {{Formal verification in hardware design: a survey}},
%journal = {ACM Trans. Des. Autom. Electron. Syst.},
year = {1999},
}

@INPROCEEDINGS{gadde,
  author={Gadde, Deepak Narayan and Radhakrishna, Keerthan Kopparam and Viswambharan, Vaisakh Naduvodi and Kumar, Aman and Lettnin, Djones and Kunz, Wolfgang and Simon, Sebastian},
  booktitle={SBCCI}, 
  title={{Hey AI, Generate Me a Hardware Code! Agentic AI-based Hardware Design \& Verification}}, 
  year={2025},
}

@book{seligman2015formal,
  title={{Formal Verification: An Essential Toolkit for Modern VLSI Design}},
  author={Seligman, E. and Schubert, T. and Kumar, M.V.A.K.},
  year={2015},
  publisher={Morgan Kaufmann}
}

@misc{nvidiacvdp,
  author    = {Nathaniel Pinckney and Chenhui Deng and Chia-Tung Ho and Yun-Da Tsai and Mingjie Liu and Wenfei Zhou and Brucek Khailany and Haoxing Ren},
  title     = {{Comprehensive Verilog Design Problems: A Next-Generation Benchmark Dataset for Evaluating Large Language Models and Agents on RTL Design and Verification}},
  howpublished={arXiv},
  year ={2025},
}

@misc{VerStudy,
	AUTHOR =        {Harry Foster},
	TITLE =         {{Wilson Research Group IC/ASIC functional verification trend report}},
	howpublished =   {Siemens Blog},
	YEAR  =         {2024}
}

@misc{meta2024llama3_3,
  author       = {{Meta AI}},
  title        = {{LLaMA 3.3 Model Card and Prompt Formats}},
  year         = {2024},
}

@misc{openai2024,
  author       = {OpenAI},
  title        = {{GPT-4.1 and GPT-5}},
  year ={2024},
}

@Article{unknown,
AUTHOR = {Abdollahi, Meisam and Yeganli, Seyedeh Faegheh and Baharloo, Mohammad (Amir) and Baniasadi, Amirali},
TITLE = {{Hardware Design and Verification with Large Language Models: A Scoping Review, Challenges, and Open Issues}},
JOURNAL = {{Electronics}},
YEAR = {2025},
}

@misc{deepseek,
      title={{DeepSeek-R1: Incentivizing Reasoning Capability in LLMs via Reinforcement Learning}}, 
      author={DeepSeek-AI and Daya Guo and Dejian Yang and Haowei Zhang and Junxiao Song and Ruoyu Zhang and Runxin Xu and Qihao Zhu and Shirong Ma and Peiyi Wang and Xiao Bi and Xiaokang Zhang and Xingkai Yu and Yu Wu and Z. F. Wu and Zhibin Gou and Zhihong Shao and Zhuoshu Li and Ziyi Gao and Aixin Liu and Bing Xue and Bingxuan Wang and Bochao Wu and Bei Feng and Chengda Lu and Chenggang Zhao and Chengqi Deng and Chenyu Zhang and Chong Ruan and Damai Dai and Deli Chen and Dongjie Ji and Erhang Li and Fangyun Lin and Fucong Dai and Fuli Luo and Guangbo Hao and Guanting Chen and Guowei Li and H. Zhang and Han Bao and Hanwei Xu and Haocheng Wang and Honghui Ding and Huajian Xin and Huazuo Gao and Hui Qu and Hui Li and Jianzhong Guo and Jiashi Li and Jiawei Wang and Jingchang Chen and Jingyang Yuan and Junjie Qiu and Junlong Li and J. L. Cai and Jiaqi Ni and Jian Liang and Jin Chen and Kai Dong and Kai Hu and Kaige Gao and Kang Guan and Kexin Huang and Kuai Yu and Lean Wang and Lecong Zhang and Liang Zhao and Litong Wang and Liyue Zhang and Lei Xu and Leyi Xia and Mingchuan Zhang and Minghua Zhang and Minghui Tang and Meng Li and Miaojun Wang and Mingming Li and Ning Tian and Panpan Huang and Peng Zhang and Qiancheng Wang and Qinyu Chen and Qiushi Du and Ruiqi Ge and Ruisong Zhang and Ruizhe Pan and Runji Wang and R. J. Chen and R. L. Jin and Ruyi Chen and Shanghao Lu and Shangyan Zhou and Shanhuang Chen and Shengfeng Ye and Shiyu Wang and Shuiping Yu and Shunfeng Zhou and Shuting Pan and S. S. Li and Shuang Zhou and Shaoqing Wu and Shengfeng Ye and Tao Yun and Tian Pei and Tianyu Sun and T. Wang and Wangding Zeng and Wanjia Zhao and Wen Liu and Wenfeng Liang and Wenjun Gao and Wenqin Yu and Wentao Zhang and W. L. Xiao and Wei An and Xiaodong Liu and Xiaohan Wang and Xiaokang Chen and Xiaotao Nie and Xin Cheng and Xin Liu and Xin Xie and Xingchao Liu and Xinyu Yang and Xinyuan Li and Xuecheng Su and Xuheng Lin and X. Q. Li and Xiangyue Jin and Xiaojin Shen and Xiaosha Chen and Xiaowen Sun and Xiaoxiang Wang and Xinnan Song and Xinyi Zhou and Xianzu Wang and Xinxia Shan and Y. K. Li and Y. Q. Wang and Y. X. Wei and Yang Zhang and Yanhong Xu and Yao Li and Yao Zhao and Yaofeng Sun and Yaohui Wang and Yi Yu and Yichao Zhang and Yifan Shi and Yiliang Xiong and Ying He and Yishi Piao and Yisong Wang and Yixuan Tan and Yiyang Ma and Yiyuan Liu and Yongqiang Guo and Yuan Ou and Yuduan Wang and Yue Gong and Yuheng Zou and Yujia He and Yunfan Xiong and Yuxiang Luo and Yuxiang You and Yuxuan Liu and Yuyang Zhou and Y. X. Zhu and Yanhong Xu and Yanping Huang and Yaohui Li and Yi Zheng and Yuchen Zhu and Yunxian Ma and Ying Tang and Yukun Zha and Yuting Yan and Z. Z. Ren and Zehui Ren and Zhangli Sha and Zhe Fu and Zhean Xu and Zhenda Xie and Zhengyan Zhang and Zhewen Hao and Zhicheng Ma and Zhigang Yan and Zhiyu Wu and Zihui Gu and Zijia Zhu and Zijun Liu and Zilin Li and Ziwei Xie and Ziyang Song and Zizheng Pan and Zhen Huang and Zhipeng Xu and Zhongyu Zhang and Zhen Zhang},
      howpublished={arXiv},
      year={2025},
}

@INPROCEEDINGS{assertllm,
  author={Fang, Wenji and Li, Mengming and Li, Min and Yan, Zhiyuan and Liu, Shang and Zhang, Hongce and Xie, Zhiyao},
  booktitle={LAD}, 
  title={{AssertLLM: Generating Hardware Verification Assertions from Design Specifications via Multi-LLMs}}, 
  howpublished={arXiv},
  year={2024},
}

@misc{MAS,
      title={{Know the Ropes: A Heuristic Strategy for LLM-based Multi-Agent System Design}}, 
      author={Zhenkun Li and Lingyao Li and Shuhang Lin and Yongfeng Zhang},
      howpublished={arXiv},
      year={2025}, 
}

@misc{saarthi,
      title={{Saarthi: The First AI Formal Verification Engineer}}, 
      author={Aman Kumar and Deepak Narayan Gadde and Keerthan Kopparam Radhakrishna and Djones Lettnin},
      howpublished={arXiv},
      year={2025},
       
}

@misc{autogen2024,
	author = {{Microsoft Research}},
	title = {{AutoGen: Framework for Multi-Agent Systems}},
	year = {2024},

}

@misc{cadenceJasper,
	author = {{Cadence Design Systems, Inc.}},
    title = {{Jasper Formal Verification Platform}},
    year={2025},
}

@misc{jeyakumar2024advancing,
title={{Advancing Agentic Systems: Dynamic Task Decomposition, Tool Integration and Evaluation using Novel Metrics and Dataset}},
author={Shankar Kumar Jeyakumar and Alaa Alameer Ahmad and Adrian Garret Gabriel},
howpublished={arXiv},
year={2024},

}

@INPROCEEDINGS{Self-ref,
  author={Renze, Matthew and Guven, Erhan},
  booktitle={FLLM}, 
  title={{Self-Reflection in Large Language Model Agents: Effects on Problem-Solving Performance}}, 
  year={2024},
  
 }

@misc{chen2025,
      title={{A Survey on LLM-based Multi-Agent System: Recent Advances and New Frontiers in Application}}, 
      author={Shuaihang Chen and Yuanxing Liu and Wei Han and Weinan Zhang and Ting Liu},
      howpublished={arXiv},
      year={2025},
}

@misc{TOOLS,
      title={{AgentOrchestra: Orchestrating Hierarchical Multi-Agent Intelligence with the Tool-Environment-Agent(TEA) Protocol}}, 
      author={Wentao Zhang and Liang Zeng and Yuzhen Xiao and Yongcong Li and Ce Cui and Yilei Zhao and Rui Hu and Yang Liu and Yahui Zhou and Bo An},
      howpublished={arXiv},
      year={2025},
}

@INPROCEEDINGS{Orch,
  author={Yu, Chaojia and Cheng, Zihan and Cui, Hanwen and Gao, Yishuo and Luo, Zexu and Wang, Yijin and Zheng, Hangbin and Zhao, Yong},
  booktitle={ICAIBD}, 
  title={{A Survey on Agent Workflow – Status and Future}},
  year={2025},
}

\end{document}